\documentclass[10pt,twocolumn,letterpaper]{article}

\usepackage[pagenumbers]{cvpr} % To force page numbers, e.g. for an arXiv version

\usepackage{url}            % simple URL typesetting
\usepackage{booktabs}       % professional-quality tables
\usepackage{amsfonts}       % blackboard math symbols
\usepackage{nicefrac}       % compact symbols for 1/2, etc.
\usepackage{microtype}      % microtypography
\usepackage{xcolor}         % colors
\usepackage{algorithm}
\usepackage{algpseudocode}
\usepackage{tabularx}
\usepackage{adjustbox}
\usepackage{multirow}
\usepackage{amssymb, amsmath, bm}
\usepackage{bbm}
\usepackage{graphicx}
\usepackage{wrapfig}
\usepackage{makecell}
\usepackage{comment}
\usepackage{array, bigdelim, makecell, enumitem}

\newcommand{\stdv}[1]{\scriptsize$\pm$#1}
\newcommand{\vm}[1]{\vspace{+0.2cm}}

\usepackage[pagebackref,breaklinks,colorlinks]{hyperref}

\usepackage[capitalize]{cleveref}
\crefname{section}{Sec.}{Secs.}
\Crefname{section}{Section}{Sections}
\Crefname{table}{Table}{Tables}
\crefname{table}{Tab.}{Tabs.}

\def\cvprPaperID{8333} % *** Enter the CVPR Paper ID here
\def\confName{CVPR}
\def\confYear{2023}

\begin{document}

\title{The Dialog Must Go On: \\ Improving Visual Dialog via Generative Self-Training}

\author{Gi-Cheon Kang$^{1,2}$ \;\; Sungdong Kim$^{3}$\thanks{Equal contribution} \;\; Jin-Hwa Kim$^{3,2*}$ \;\; Donghyun Kwak$^{4*}$ \;\; Byoung-Tak Zhang$^{1,2}$\\
  $^1$IPAI, Seoul National University \;\;\;\; $^2$AIIS \;\;\;\; $^3$NAVER AI Lab \;\;\;\; $^4$NAVER Cloud CLOVA \\
  {\tt\small \{chonkang, btzhang\}@snu.ac.kr \, \{sungdong.kim, j1nhwa.kim, donghyun.kwak\}@navercorp.com} 
%\and
}
\maketitle

%%%%%%%%% ABSTRACT
\begin{abstract}
   Visual dialog (VisDial) is a task of answering a sequence of questions grounded in an image, using the dialog history as context. Prior work has trained the dialog agents solely on VisDial data via supervised learning or leveraged pre-training on related vision-and-language datasets. This paper presents a semi-supervised learning approach for visually-grounded dialog, called \textit{Generative Self-Training} (GST), to leverage unlabeled images on the Web. Specifically, GST first retrieves in-domain images through out-of-distribution detection and generates synthetic dialogs regarding the images via multimodal conditional text generation. GST then trains a dialog agent on the synthetic and the original VisDial data. As a result, GST scales the amount of training data up to an order of magnitude that of VisDial (1.2M $\rightarrow$ 12.9M QA data). For robust training of the synthetic dialogs, we also propose perplexity-based data selection and multimodal consistency regularization. Evaluation on VisDial v1.0 and v0.9 datasets shows that GST achieves new state-of-the-art results on both datasets. We further observe the robustness of GST against both visual and textual adversarial attacks. Finally, GST yields strong performance gains in the low-data regime. Code is available at \url{https://github.com/gicheonkang/gst-visdial}.
\end{abstract}

\section{Introduction}
\label{sec:intro}

% paragraph 1
% task definition why this task is important, which aspect should be considered to deal with this challenge.
Recently, there has been extensive research towards developing visually-grounded dialog systems \cite{das2017visual, de2017guesswhat,Kottur2019CLEVRDialog,kim2017codraw} due to their significance in many real-world applications (\textit{e.g.,} helping visually impaired person). Notably, Visual Dialog (VisDial)~\cite{das2017visual} has provided a testbed for studying such systems, where a dialog agent should answer a \textit{sequence} of image-grounded questions. For instance, the agent is expected to answer open-ended questions like \textit{``What color is it?''} and \textit{``How old does she look?''}. This task requires a 
holistic understanding of visual information, linguistic semantics in context (\textit{e.g.,} it and she), and most importantly, the grounding of these two.

% paragraph 2
% prevailing research trend in VisDial: pretrain-then-transfer
Most of the previous approaches in VisDial~\cite{lu2017best,seo2017visual,wu2018you,kottur2018visual,niu2018recursive,schwartz2019factor,guo2019image,gan2019multi,kang2019dual,zheng2019reasoning,chen2020dmrm,jiang2020dam,jiang2020kbgn,nguyen2020efficient,chen2021multimodal,kang2020reasoning} have trained the dialog agents solely on VisDial data via supervised learning. More recent studies \cite{murahari2019large,wang2020vd,chen2022utc} have employed self-supervised pre-trained models such as BERT~\cite{devlin2018bert} or ViLBERT~\cite{lu2019vilbert} and finetuned them on VisDial data. The models are typically pre-trained to recover masked inputs and predict the semantic alignment between two segments. This \textit{pretrain-then-transfer} learning strategy has shown promising results by transferring knowledge from the models pre-trained on large-scale data sources~\cite{sharma2018conceptual,antol2015vqa,zhu2015book} to VisDial.

% paragraph 3 
% Considerations for applying SSL to VisDial
Our research question is the following: \textit{How can the dialog agent expand its knowledge beyond what it can acquire via supervised learning or self-supervised pre-training on the provided datasets?} Some recent studies have shown that semi-supervised learning and pre-training have complementary modeling capabilities in image~\cite{zoph2020rethinking} and text classification~\cite{du2020self}. Inspired by them, we consider semi-supervised learning (SSL) as a way to address the above question. 

Let us assume that large amounts of unlabeled images are available. SSL for VisDial can be applied to generate synthetic conversations for the unlabeled images and train the agent with the synthetic data. However, there are two critical challenges to this approach. First, the target output for VisDial (\textit{i.e.,} multi-turn visual QA data) is more complex than that of the aforementioned studies~\cite{zoph2020rethinking,du2020self}. Specifically, they have addressed the classification problems, yielding class probabilities as pseudo labels~\cite{lee2013pseudo}. In contrast, SSL for VisDial should generate a sequence of pseudo queries (\textit{i.e.,} visual questions) and pseudo labels (\textit{i.e.,} corresponding answers) in \textit{natural language} to train the answering agent. It further indicates that the target output should be generated while considering the \textit{multimodal} and \textit{sequential} nature of the visual dialog task. Next, even if SSL yields synthetic dialogs via text generation, there may be noise, such as generating irrelevant questions or incorrect answers to given contexts. A robust training method is required to leverage such noisy synthetic dialog datasets.

% paragraph
In this paper, we study the above challenges in the context of SSL, especially self-training~\cite{zoph2020rethinking,du2020self,lee2013pseudo,berthelot2019mixmatch,sohn2020fixmatch,xie2020unsupervised,xie2020self,he2019revisiting,scudder1965probability,li2019learning,rosenberg2005semi,mukherjee2020uncertainty,karamanolakis2021self,jo2019delta,thakur2020augmented}, where a teacher model trained on labeled data predicts the pseudo labels for unlabeled data. Then, a student model jointly learns on the labeled and the pseudo-labeled datasets. Unlike existing studies in self-training that have mainly studied uni-modal, discriminative tasks such as image classification~\cite{xie2020self,zoph2020rethinking,sohn2020fixmatch} or text classification~\cite{du2020self,mukherjee2020uncertainty,karamanolakis2021self}, we extend the idea of self-training to the task of multimodal conditional text generation. 

To this end, we propose a new learning strategy, called \textit{Generative Self-Training} (GST), that artificially generates multi-turn visual QA data and utilizes the synthetic data for training. GST first trains the teacher model (answerer) and the visual question generation model (questioner) using VisDial data. It then retrieves a set of unlabeled images from a Web image dataset, Conceptual 12M~\cite{changpinyo2021conceptual}. Next, the questioner and the teacher generate a series of visual QA pairs for the retrieved images. Finally, the student is trained on the synthetic and the original VisDial data. We also propose perplexity-based data selection (PPL) and multimodal consistency regularization (MCR) to effectively train the student with the noisy dialog data. PPL is to selectively utilize the answers whose perplexity of the teacher is below a threshold. MCR encourages the student to yield consistent predictions when the perturbed multimodal inputs are given. As a result, GST successfully augments the synthetic VisDial data (11.7M QA pairs), thus mitigating the need to scale up the size of the human-annotated VisDial data, which is prohibitively expensive and time-consuming. 

Our key contributions are three-fold. First, we propose Generative Self-Training (GST) that generates multi-turn visual QA data to leverage unlabeled Web images effectively. Second, experiments show that GST achieves new state-of-the-art performance on VisDial v1.0 and v0.9 datasets. We further demonstrate two important results: (1) GST is indeed effective when the human-annotated visual dialog data is extremely scarce (improving up to 11.09 absolute points on NDCG), and (2) PPL and MCR are effective when training the noisy synthetic dialog data. Third, to validate the robustness of GST, we evaluate our proposed method under three different visual and textual adversarial attacks, \ie, FGSM, coreference, and random token attacks. We observe that GST significantly improves the performance compared with the baseline models against all adversarial attacks, especially boosting NDCG scores from 21.60\% to 45.43\% in the FGSM attack~\cite{goodfellow2014explaining}.

%\begin{itemize}[leftmargin=0.4cm]
%\item[$\bullet$] We propose Generative Self-Training (GST) that generates multi-turn visual QA data to leverage unlabeled Web images effectively. %To the best of our knowledge, GST is the first semi-supervised learning approach throughout a wide range of visual QA tasks, including VisDial.
%\item[$\bullet$] Experiments show that GST achieves the new state-of-the-art performance on VisDial v1.0 and v0.9 datasets. We further demonstrate two important results: (1) GST is indeed effective in the low-data regime, and (2) the perplexity-based data selection (PPL) and the multimodal consistency regularization (MCR) are effective when training the noisy synthetic dialog data. 
%\item[$\bullet$] \rev{We introduce a new evaluation setup for adversarial robustness in VisDial and then evaluate GST in that setting. GST significantly improves the robustness against three different visual and textual adversarial attacks, \ie, FGSM, dialog round, and random token attacks.}
%\end{itemize}
\section{Related work}
\label{sec:related_work}
\noindent\textbf{Visual dialog.} Visual Dialog (VisDial)~\cite{das2017visual} has been proposed as an extended version of Visual Question Answering (VQA)~\cite{antol2015vqa,Anderson2017up-down,kim2018bilinear}, where a dialog agent should answer a series of interdependent questions using an image and the dialog history. Prior work has developed a variety attention mechanisms~\cite{lu2017best,seo2017visual,wu2018you,kottur2018visual,niu2018recursive,schwartz2019factor,guo2019image,gan2019multi,kang2019dual,nguyen2020efficient} considering the interactions among the image, dialog history, and question. Some studies~\cite{zheng2019reasoning,kang2020reasoning} have attempted to discover the semantic structures of the dialog in the context of graph neural networks~\cite{scarselli2008graph} using the soft attention mechanisms~\cite{bahdanau2014neural}. From the learning algorithm perspective, all of them have relied on supervised learning on VisDial data. More recently, a line of research~\cite{murahari2019large,wang2020vd,chen2022utc} has employed self-supervised pre-training to leverage the knowledge of related vision-and-language datasets~\cite{sharma2018conceptual,antol2015vqa,zhu2015book}. However, our approach is based on semi-supervised learning and produces the task-specific data (\textit{i.e.,} visual dialogs) for unlabeled images to train the dialog agent. \\

\noindent\textbf{Sequence generation in vision-and-language tasks.} Many studies have generated natural language for the visual inputs such as image captioning~\cite{xu2015show,Anderson2017up-down}, video captioning~\cite{iashin2020multi,pan2017video}, visual question generation (VQG)~\cite{kai2021learning,krishna2019information,fan2018question,liu2018ivqa,patro2018multimodal,jain2017creativity}, visual dialog (VisDial)~\cite{das2017visual,gan2019multi}, and video dialog~\cite{alamri2019audio,le2019multimodal}. Furthermore, recent studies~\cite{yang2021just,li2022blip} have produced text data for vision-and-language pre-training. GST is similar to these studies in that the model generates the text data, but our focus is on studying the effect of semi-supervised learning (SSL) on top of such pre-training approaches. To the best of our knowledge, GST is the first approach to show the efficacy of SSL throughout a wide range of visual QA tasks. \\

\noindent\textbf{Neural dialog generation.} In NLP literature, extensive studies have been conducted regarding neural dialogue generation for both open-domain dialogue~\cite{zhang2019dialogpt,shang2015neural,li2016deep,serban2017hierarchical,saleh2020hierarchical,li2017adversarial} and task-oriented dialogue~\cite{wang2020multi,huang2020mala}. Our approach is similar to neural dialogue generation in that the model should generate a corresponding response based on the dialog history and the current utterance. However, we aim to produce \textit{visually-grounded} dialogs, and thus the image-groundedness of the question and the semantic correctness of the answer are important. On the other hand, neural dialogue generation considers many different aspects: specificity, response-relatedness~\cite{see2019makes}, interestingness~\cite{mehri2020unsupervised}, and diversity~\cite{li2016deep}.

\section{Approach}
\label{sec:approach}
%In this section, we begin by formally describing self-training and the visual dialog task (Sec. \hyperref[sec:preliminaries]{3.1}). We then describe how we extend self-training to our proposed method for VisDial (Sec. \hyperref[sec:gst]{3.2}).   
\begin{figure*}[t!]
\centering
\label{fig:teaser}
\includegraphics[width=\textwidth]{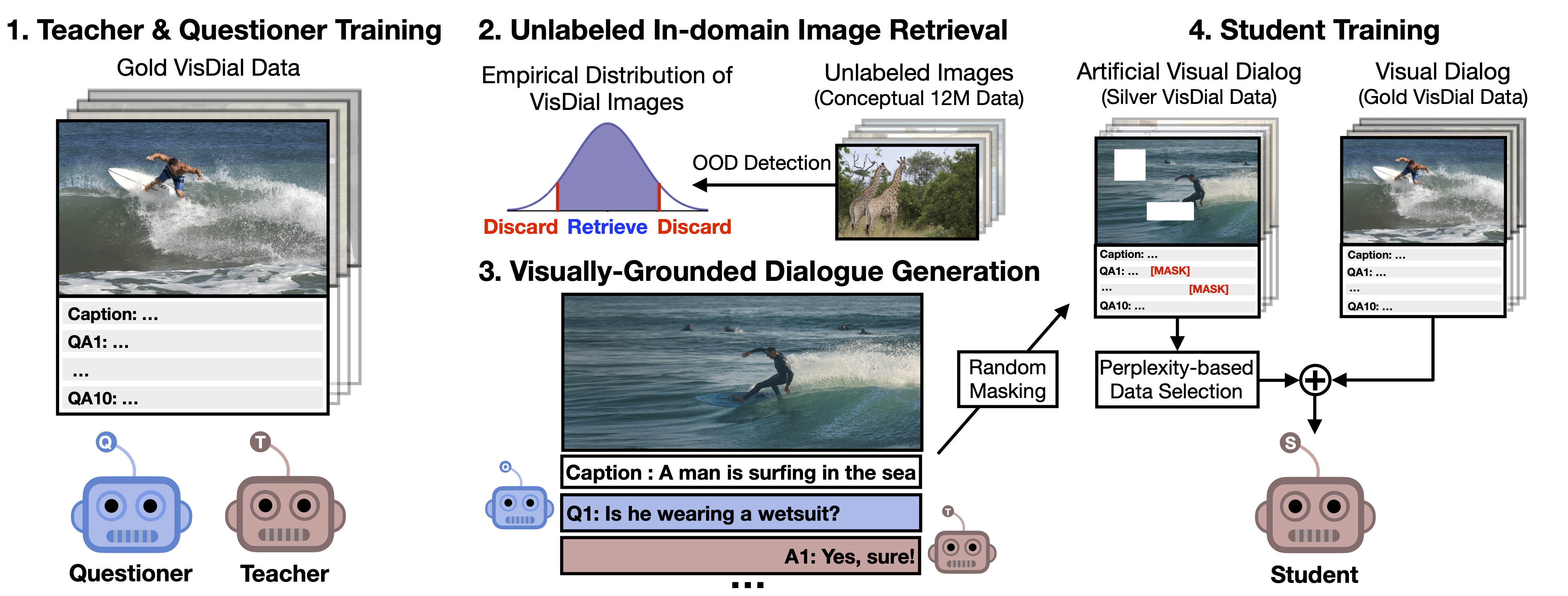}
\caption{An overview of Generative Self-Training (GST).}
\end{figure*}

\subsection{Preliminaries}
\label{sec:preliminaries}
\noindent\textbf{Self-training.} We have a labeled dataset $L$ = $\{(x_n, y_n)\}_{n=1}^N$ and an unlabeled dataset $U$ = $\{\tilde{x}_m\}_{m=1}^M$. Typically, self-training trains a teacher model $P_{\mathcal{T}}$ on the labeled dataset $L$. The teacher then predicts the pseudo label $\tilde{y}$ for the unlabeled data $\tilde{x} \sim U$, constructing the pseudo-labeled dataset $\tilde{L}$ = $\{(\tilde{x}_m, \tilde{y}_m)\}_{m=1}^M$. Finally, a student model $P_{\mathcal{S}}$ is trained on $L \cup \tilde{L}$. Many variants have been studied on this setup: (1) selecting the subset of the pseudo-labeled dataset \cite{he2019revisiting,xie2020self,sohn2020fixmatch}, (2) adding noise to inputs \cite{zoph2020rethinking,he2019revisiting,xie2020self,xie2020unsupervised,sohn2020fixmatch}, and (3) iterating the above setup multiple times \cite{he2019revisiting,xie2020self}. \\

\noindent\textbf{Visual dialog.} The visual dialog (VisDial) dataset \cite{das2017visual} contains an image $v$ and a visually-grounded dialog $d$ = $\{\underbrace{c}_{d_{0}}, \underbrace{(q_{1}, a^{gt}_{1})}_{d_{1}}, \cdots, \underbrace{(q_{T}, a^{gt}_{T})}_{d_{T}}\}$ where $c$ denotes an image caption. $T$ is the number of rounds for each dialog. At round $t$, a dialog agent is given a triplet $(v,d_{<t}, q_t)$ as an input, consisting of the image, the dialog history, and a visual question. $d_{<t}$ denotes all dialog rounds before the $t$-th round. The agent is then expected to predict a ground-truth answer $a^{gt}_t$. There are two broad classes of methods in VisDial: \textit{generative} and \textit{discriminative}. Generative models aim to generate the ground-truth answer by maximizing the log-likelihood of $a^{gt}_t$. In contrast, discriminative models are trained to retrieve the ground-truth answer from a list of answer candidates {\it $a^{gt}_t$ $\in$ $\left\{a_{t}^1, \cdots, a_{t}^{100} \right\}$}. Our main focus is the generative models since they do not need pre-defined answer candidates and are thus more practical to be deployed in real-world applications. 

\subsection{Generative Self-Training (GST)}
\label{sec:gst}
This subsection describes our approach, called GST, which generates multi-turn visual QA data and utilizes the generated data for training. An overview of GST is shown in Figure~\hyperref[fig:teaser]{1}. We have a human-labeled VisDial dataset $L = \{ (v_n, d_n) \}_{n=1}^N$ where $v_n$ is a given image, and each dialog $d_n = \{\underbrace{c_n}_{d_{n, 0}}, \underbrace{(q_{n,1}, a^{gt}_{n,1})}_{d_{n, 1}}, \cdots, \underbrace{(q_{n,T}, a^{gt}_{n,T})}_{d_{n,T}}\}$ consists of an image caption $c$ and $T$ rounds of QA pairs. In the following, we omit the superscript $gt$ in the ground-truth answer for brevity. GST first trains a teacher $P_{\mathcal{T}}$ and a questioner $P_{\mathcal{Q}}$ with the labeled dataset $L$ via supervised learning. It then retrieves unlabeled images $U = \{ \tilde{v}_m \}_{m=1}^M$ from the Conceptual 12M dataset~\cite{changpinyo2021conceptual} using a simple outlier detection model, the multivariate normal distribution. Next, the questioner and the teacher generate the visually-grounded dialog $\tilde{d}$ for the unlabeled image $\tilde{v}$ via multimodal conditional text generation, finally yielding a synthetic dialog dataset $\tilde{L} = \{ (\tilde{v}_m, \tilde{d}_m) \}_{m=1}^M$. We call this dataset the \textit{silver VisDial} data to distinguish it from the human-labeled VisDial dataset~\cite{das2017visual} (short for the \textit{gold VisDial} data). Finally, a student $P_{\mathcal{S}}$ is trained on a combination of the gold and the silver VisDial data while applying perplexity-based data selection (PPL) and multimodal consistency regularization (MCR) to the silver VisDial data. We describe the details of each process in the following parts. \\   

\noindent\textbf{Teacher \& questioner training.} First, a series of question-and-answer pairs for the unlabeled images should be generated to train the answering agent. Accordingly, GST first trains the answer generator, the teacher model $P_{\mathcal{T}}$, on the gold VisDial dataset. Specifically, the teacher learns to generate the ground-truth answer's word sequence $a_{t} = (w_{t,1}, \cdots, w_{t,S})$, given the context triplet $c_{t} \triangleq (v, d_{<t}, q_{t})$, consisting of the image, the dialog history, and the question. It is optimized by minimizing the negative log-likelihood of the ground-truth answer. Formally, 
\begin{equation}
\begin{split}
    \mathcal{L}_{\mathcal{T}} &= -{1 \over NT} \sum_{n=1}^N \sum_{t=1}^T \mathrm{log}\,P_\mathcal{T}(a_{n, t} | c_{n, t}) \\
    &= -{1 \over NTS} \sum_{n=1}^N \sum_{t=1}^T \sum_{s=1}^S \mathrm{log}\,P_\mathcal{T}(w_s | c_{n, t}, w_{<s})
\end{split}    
\end{equation}
where $N$, $T$, and $S$ denote the number of data tuples in gold VisDial data, dialog rounds, and the sequence length of the ground-truth answer, respectively. $w_{<s}$ indicates all word tokens before the $s$-th token in the answer sequence. Similar to the teacher, the questioner is trained to generate the question at round $t$, given the image and the dialog history until round $t-1$ (\textit{i.e.,} $P_{\mathcal{Q}}(q_{t}|v, d_{<t})$). The questioner is also optimized by minimizing the negative log-likelihood of the follow-up question. Note that the teacher and the questioner are trained separately to prevent possible unintended co-adaptation~\cite{kim2017codraw}. Both the teacher and the questioner are based on encoder-decoder architecture, where an encoder aggregates the context triplet, and a decoder generates the target sentence. We implement the models by integrating a pre-trained vision-and-language encoder, ViLBERT~\cite{lu2019vilbert}, with the transformer decoder~\cite{rothe2020leveraging}. We refer readers to Appendix A for a detailed architecture. \\

\noindent\textbf{Unlabeled in-domain image retrieval (IIR).} Inspired by the work~\cite{du2020self} that highlighted the importance of using in-domain data, GST retrieves in-domain image data from the Conceptual 12M dataset~\cite{changpinyo2021conceptual} with an out-of-distribution (OOD) detection model. Specifically, we extract the $D$ dimensional feature vector for each image in the gold VisDial dataset by using the Vision Transformer (ViT)~\cite{dosovitskiy2020image} in the CLIP model~\cite{radford2021learning}, yielding a feature matrix for the entire images $\mathbf{X}=(X_1, \cdots, X_N)^\top \in \mathbb{R}^{N \times D}$. Based on the matrix, we build the multivariate normal distribution whose dimension is $D$, \textit{i.e.,} $\mathbf{X} \sim \mathcal{N}_D(\mu, \Sigma)$. We regard this normal distribution as the empirical distribution of the gold VisDial images and perform OOD detection by identifying the probability of each feature vector for the unlabeled image. Consequently, the top-$M$ unlabeled images are retrieved out of 12 million Web images ($M \approx 3.6$ million). \\

\noindent\textbf{Visually-grounded dialog generation.} This step mimics a scenario where two people have a conversation about the given images. Given the retrieved images $U = \{ \tilde{v}_m \}_{m=1}^M$, our goal is to generate the visually-grounded dialogs $\{\tilde{d}_m\}_{m=1}^M$ where each dialog $\tilde{d}$ consists of the image caption and $T$ rounds of QA pairs. In an actual implementation, we use the image captions in the Conceptual 12M dataset~\cite{changpinyo2021conceptual} and thus do not generate the captions. The QA pairs are sequentially generated. Concretely, the image $\tilde{v}$, the caption $\tilde{c}$, and the generated QA pairs until round $t-1$ are used as inputs when the questioner generates the question at round $t$ (\textit{i.e.,} $\tilde{q}_t$). After then, the teacher produces the corresponding answer $\tilde{a}_t$ based on the image $\tilde{v}$, the dialog history $\tilde{d}_{<t}$, and the question $\tilde{q}_t$. Finally, GST produces the silver VisDial dataset $\tilde{L} = \{ (\tilde{v}_m, \tilde{d}_m) \}_{m=1}^M$. \\

\noindent\textbf{Student training with noisy data.} In Figure~\hyperref[fig:teaser]{1}, the student $P_\mathcal{S}$ is trained on the combination of the silver and the gold VisDial data. According to many studies~\cite{xie2020self,he2019revisiting,sohn2020fixmatch,zoph2020rethinking} in self-training, selectively utilizing the samples in the pseudo-labeled dataset is a common approach since the confidence of the teacher model's predictions varies from sample to sample. To this end, we introduce a simple yet effective data selection method for the sequence generation problem, perplexity-based data selection (PPL). PPL is to utilize the answers whose perplexity of the teacher is below a certain threshold. Perplexity is defined as the exponentiated average negative log-likelihood of a sequence; the lower, the better. We hypothesize that PPL, albeit noisy, can be an indicator of whether the generated answer is correct or not, as in~\cite{shakeri2020end}. Furthermore, inspired by the consistency regularization~\cite{xie2020unsupervised,sohn2020fixmatch} widely utilized in recent SSL algorithms, we also propose the multimodal consistency regularization (MCR) to improve the generalization capability of the student. MCR encourages the student to yield predictions similar to the teacher's predictions even when the student is provided with perturbed multimodal inputs. Finally, we design a loss function for the student as:
\begin{equation}
\small
\begin{split}
    &\mathcal{L}_{\mathcal{S}} =\\
    &~-{1 \over MT} \sum_{m=1}^M \sum_{t=1}^T \mathbbm{1}(\mathrm{PPL}(\tilde{a}_{m, t})<\tau)  \mathrm{log}\underbrace{P_\mathcal{S}(\tilde{a}_{m, t} | \mathcal{M}(\tilde{c}_{m, t}))}_{\mathrm{MCR}} \\
    &~-{1 \over NT} \sum_{n=1}^N \sum_{t=1}^T \mathrm{log} P_\mathcal{S}(a_{n, t} | c_{n, t}) \\
    &~\mathrm{where} \,\,\, \mathrm{PPL}(\tilde{a}_t) = \mathrm{exp}\left\{{-{1 \over S}} \sum_{s=1}^S \mathrm{log} P_\mathcal{T}(\tilde{w}_s | \tilde{c}_{t}, \tilde{w}_{<s})\right\}
\end{split}
\end{equation}
where $M$, $\mathbbm{1}$, and $\tau$ denote the number of data tuples in silver VisDial data, indicator function, and selection threshold, respectively. $\tilde{c}_{m,t} \triangleq (\tilde{v}_m, \tilde{d}_{m,<t}, \tilde{q}_{m,t})$ denotes the context for the silver VisDial data. The loss function is the sum of the losses for the silver and the gold VisDial data. PPL and MCR are applied to compute the loss of the silver VisDial data. PPL is used in the indicator function above, selecting the synthetic answers whose perplexity of the teacher is below the threshold $\tau$. It implies that the unselected answers are ignored during training. The teacher's perplexity of each answer is computed in the dialog generation step above. Next, $\mathcal{M}$ denotes the stochastic function for MCR that injects perturbations to the input space of the student. Inspired by ViLBERT~\cite{lu2019vilbert}, we implement the stochastic function by randomly masking 15\% of image regions and word tokens. Specifically, masked image regions have their image features zeroed out, and the masked word tokens are replaced with a special \texttt{[MASK]} token. The intuition behind MCR is minimizing the distance between the \textit{perturbed} (\textit{i.e.,} masked) predictions from the student and the \textit{unperturbed} predictions (\textit{i.e.,} $\tilde{a}_{m, t}$) from the teacher. It indicates that the perturbation is not injected when the teacher generates the synthetic answers. We believe MCR makes the student robust to the input noise, and PPL encourages the student to maintain a low entropy (\textit{i.e.,} confident) in noisy data training. The student and the teacher have the same model capacity and are based on the same model architecture.

\section{Experiments}

\subsection{Experimental setup}
\label{sec:setup}

\noindent\textbf{{VisDial datasets.}} We evaluate our proposed approach on the VisDial v1.0 and v0.9 datasets~\cite{das2017visual}, collected by the AMT chatting between two workers about MS-COCO~\cite{lin2014microsoft} images. Each dialog consists of a caption from COCO and a sequence of ten QA pairs. The VisDial v0.9 dataset has 83k dialogs on COCO-train and 40k dialogs on COCO-validation images. More recently, Das \textit{et al.} \cite{das2017visual} released additional 10k dialogs on Flickr images to use them as validation and test splits for the VisDial v1.0 dataset. As a result, the VisDial v1.0 dataset contains 123k, 2k, and 8k dialogs as train, validation, and test split. This dataset is licensed under a Creative Commons Attribution 4.0 International License. \\

\noindent\textbf{{Evaluation protocol.}} We follow the standard evaluation protocol established in the work~\cite{das2017visual} for evaluating visual dialog models. The visual dialog models for both generative and discriminative tasks have been evaluated by the retrieval-based evaluation metrics: mean reciprocal rank (MRR), recall@k (R@k), mean rank (Mean), and normalized discounted cumulative gain (NDCG). Specifically, all dialogs in VisDial contain a list of 100 answer candidates for each visual question, and there is one ground-truth answer in the answer candidates. The model sorts the answer candidates by the log-likelihood scores and then is evaluated by the four different metrics. MRR, R@k, and Mean consider the rank of the single ground-truth answer, while NDCG\footnote{https://visualdialog.org/challenge/2019\#evaluation} considers all relevant answers from the 100-answers list by using the densely annotated relevance scores for all answer candidates. The community regards NDCG as the primary evaluation metric. \\

\noindent\textbf{{The size of synthetic data.}} The size of the silver VisDial data (\textit{i.e.,} $M$) is 3.6M which is 30x larger than that of the gold VisDial data ($N=0.12$M). Note that the silver VisDial data contains approximately 36M QA pairs since each dialog contains 10 QA pairs. 11.7M QA pairs out of 36M ($\sim$32\%) are actually utilized after applying perplexity-based data selection when we set the selection threshold $\tau$ to 50. Consequently, the total amount of the training data is nearly 12.9M QA pairs, combining the silver data (11.7M QA pairs) with the original gold data (1.2M QA pairs). \\

\noindent\textbf{{Iterative training.}} We introduce the concept of iterative training \cite{xie2020self,he2019revisiting}, which iterates the self-training algorithm a few times. The iterative training treats the student model at $i$-th iteration as a teacher model at ($i$+1)-th iteration to generate a new synthetic silver data and train a new student. Specifically, the iterative training repeats the third and fourth steps in Figure~\hyperref[fig:teaser]{1}, where the silver VisDial data accumulates as the iteration proceeds. The student model at each iteration is trained with the accumulated silver and gold data by following the previous studies~\cite{xie2020self,he2019revisiting}. We iterate GST up to three times. Unless stated otherwise, the student model is trained with three iterations. 

\subsection{Visual dialog results}
\label{sec:quan}
\noindent\textbf{Comparison with state-of-the-art.} We compare GST with the state-of-the-art approaches on the validation set of the VisDial v1.0 and v0.9 datasets, consisting of UTC~\cite{chen2022utc}, MITVG~\cite{chen2021multimodal}, VD-BERT~\cite{wang2020vd}, LTMI~\cite{nguyen2020efficient}, KBGN~\cite{jiang2020kbgn}, DAM~\cite{jiang2020dam}, ReDAN~\cite{gan2019multi}, DMRM~\cite{chen2020dmrm}, Primary~\cite{guo2019image}, RvA~\cite{niu2018recursive}, CorefNMN~\cite{kottur2018visual}, CoAtt~\cite{wu2018you}, HCIAE~\cite{lu2017best}, and MN~\cite{das2017visual}. We decided to use the validation splits since all previous studies benchmarked the models on those splits. In Table~\hyperref[tab:sota]{1}, GST significantly outperforms all compared methods on all evaluation metrics. Compared with the state-of-the-art model, the student model improves MRR 3.20\% (56.83 $\rightarrow$ 60.03) and R@1 3.26\% (47.14 $\rightarrow$ 50.40) on the VisDial v0.9 dataset. The improvement is consistently observed on the VisDial v1.0 dataset, boosting NDCG 1.61\% (63.86 $\rightarrow$ 65.47) and MRR 0.97\% (52.22 $\rightarrow$ 53.19). Moreover, it is noticeable that recent strong models (\textit{i.e.,} UTC, MITVG, and VD-BERT) are also built based on the pre-trained weights of ViLBERT~\cite{lu2019vilbert}, transformer~\cite{vaswani2017attention}, and BERT~\cite{devlin2018bert}, respectively. Our proposed method also achieves new state-of-the-art results on the discriminative VisDial models. Details can be found in Appendix B. \\

\begin{table*}[ht!]
  \centering
  \resizebox{0.95\textwidth}{!}{
  \begin{tabular}{lccccccccccccc}
    \hline
    \toprule
    \multirow{2}{*}{} & \multicolumn{5}{c}{VisDial v0.9 (val)}  & \multicolumn{6}{c}{VisDial v1.0 (val)} \\ 
    \cmidrule(lr){2-6}\cmidrule(lr){7-12}
    Model & MRR$\uparrow$ & R@1$\uparrow$ & R@5$\uparrow$ & R@10$\uparrow$ & Mean$\downarrow$ & NDCG$\uparrow$ & MRR$\uparrow$ & R@1$\uparrow$ & R@5$\uparrow$ & R@10$\uparrow$ & Mean$\downarrow$\\
    \midrule
    MN$\dagger$~\cite{das2017visual}   & 52.59 & 42.29 & 62.85 & 68.88 & 17.06 & 51.86 & 47.99 & 38.18 & 57.54 & 64.32 & 18.60  \\
    HCIAE$\dagger$~\cite{lu2017best}  & 53.86 & 44.06 & 63.55 & 69.24 & 16.01 & 59.70 & 49.07 & 39.72 & 58.23 & 64.73 & 18.43   \\
    CoAtt$\dagger$~\cite{wu2018you}  & 55.78 & 46.10 & 65.69 & 71.74 & 14.43 & 59.24 & 49.64 & 40.09 & 59.37 & 65.92 & 17.86  \\
    CorefNMN~\cite{kottur2018visual} & 53.50 & 43.66 & 63.54 & 69.93 & 15.69 & - & - & - & - & - & - \\    
    RvA~\cite{niu2018recursive} & 55.43 & 45.37 & 65.27 & 72.97 & \textbf{10.71} & - & - & - & - & - & - \\
    Primary~\cite{guo2019image}  & - & - & - & - & - & - & 49.01 & 38.54 & 59.82 & 66.94 & 16.60 \\
    DMRM~\cite{chen2020dmrm} & 55.96 & 46.20 & 66.02 & 72.43 & 13.15 & - & 50.16 & 40.15 & 60.02 & 67.21 & 15.19  \\
    ReDAN~\cite{gan2019multi} & - & - & - & - & - & 60.47 & 50.02 & 40.27 & 59.93 & 66.78 & 17.40 \\
    DAM~\cite{jiang2020dam} & - & - & - & - & - & 60.93 & 50.51 & 40.53 & 60.84 & 67.94 & 16.65 \\
    KBGN~\cite{jiang2020kbgn}  & - & - & - & - & - & 60.42 & 50.05 & 40.40 & 60.11 & 66.82 & 17.54 \\
    LTMI~\cite{nguyen2020efficient}  & - & - & - & - & - & 63.58 & 50.74 & 40.44 & 61.61 & 69.71 & 14.93 \\
    VD-BERT~\cite{wang2020vd} & 55.95 & 46.83 & 65.43 & 72.05 & 13.18 & - & - & - & - & - & - \\
    MITVG~\cite{chen2021multimodal} & \underline{56.83} & \underline{47.14} & \underline{67.19} & \underline{73.72} & \underline{11.95} & 61.47 & 51.14 & 41.03 & 61.25 & 68.49 & \underline{14.37}  \\
    UTC~\cite{chen2022utc} & - & - & - & - & - & \underline{63.86} & \underline{52.22} & \underline{42.56} & \underline{62.40} & \underline{69.51} & 15.67 \\
    \midrule
    %Teacher (iter1) & 58.08 & 48.53 & 68.84 & 75.79 & 12.88 & 64.50 & 52.06 & 42.04& 62.92 & 71.06 & 14.54 \\
    %Student (iter1) & 59.61\stdv{.10} & 50.04\stdv{.05} & 70.18\stdv{.08} & \underline{76.51}\stdv{.21} & 12.49\stdv{.20} & \underline{65.06}\stdv{.12} & 52.84\stdv{.08} & 42.74\stdv{.09} & \underline{63.66}\stdv{.10} & \underline{71.30}\stdv{.15} & 14.60\stdv{.17} \\
    %Student (iter2) & \underline{59.72}\stdv{.13} & \underline{50.12}\stdv{.08} & \underline{70.52}\stdv{.11} & 76.47\stdv{.17} & 12.38\stdv{.21} & \textbf{65.47}\stdv{.16} & \underline{53.04}\stdv{.10} &\textbf{43.15}\stdv{.09} & 63.63\stdv{.11} & 71.00\stdv{.15} & 14.73\stdv{.15} \\
    \textbf{Student (ours)} & \textbf{60.03}\stdv{.18} & \textbf{50.40}\stdv{.15} & \textbf{70.74}\stdv{.09} & \textbf{77.15}\stdv{.13} & 12.13\stdv{.18} & \textbf{65.47}\stdv{.14} & \textbf{53.19}\stdv{.11} & \textbf{43.08}\stdv{.10} & \textbf{64.09}\stdv{.05} & \textbf{71.51}\stdv{.13} & \textbf{14.34}\stdv{.15} \\
    \bottomrule
    \hline
  \end{tabular}}
  \caption{Comparison with the state-of-the-art generative models on both VisDial v0.9 and v1.0 validation datasets. $\uparrow$ indicates higher is better. $\downarrow$ indicates lower is better. NDCG is not supported in v0.9 dataset. $\dagger$ denotes that the models are re-implemented by the previous work~\cite{gan2019multi}. The standard deviations of our proposed models are reported $\pm$ with three different initialized models.}
  \label{tab:sota}
\end{table*}
\begin{table} %[ht!]
  \centering
  \resizebox{0.4\textwidth}{!}{
  \begin{tabular}{lccccc}
    \hline
    \toprule
    \multirow{2}{*}{} & \multicolumn{5}{c}{NDCG} \\ 
    \cmidrule(lr){2-6} 
    Model & 1\% & 5\% & 10\% & 20\% & 30\% \\
    \midrule
    Teacher & 27.64  & 50.04 & 54.46 & 57.14 & 60.67 \\
    \midrule
    \textbf{Student} & \textbf{\makecell{38.73 \\ (\textcolor{blue}{+11.09})}} & \textbf{\makecell{56.60 \\ (\textcolor{blue}{+6.56})}} & \textbf{\makecell{58.62 \\ (\textcolor{blue}{+4.16})}} & \textbf{\makecell{60.92 \\ (\textcolor{blue}{+3.78})}} & \textbf{\makecell{63.09 \\ (\textcolor{blue}{+2.42})}} \\
    \bottomrule
    \hline
  \end{tabular}}
  \caption{Results of GST in the low-data regime. We report NDCG scores based on the VisDial v1.0 validation split. We assume a small subset of the gold VisDial data ($\sim$30\%) is available.}
  \label{tab:low}
\end{table}

\noindent\textbf{GST in the low-data regime.} Is GST also helpful when gold data is scarce? We investigate this question to identify the effect of GST in the low-data regime. We assume that only a small subset of the gold VisDial data (1\%, 5\%, 10\%, 20\%, and 30\%) is available. Therefore, the size of the gold data is 0.01$N$, 0.05$N$, 0.1$N$, 0.2$N$, and 0.3$N$, respectively. We first train the teacher and the questioner on such scarce data, and then these two agents generate a new silver VisDial data for unlabeled images in the Conceptual 12M dataset~\cite{changpinyo2021conceptual} with size $5N$. The student is then trained on the newly generated silver VisDial data and the small amount of the gold VisDial data. The student is based on a single iterative training, and PPL and MCR are still applied in this experiment. In Table~\hyperref[tab:low]{2}, GST yields huge improvements on both metrics, especially NDCG, boosting up to 11.09 absolute points compared with the teacher. We observe that the smaller the amount of gold data, the larger the performance gap between the teacher and the student on NDCG. It implies that GST is helpful, especially when gold data is scarce. We speculate the results in the low-data regime are particularly remarkable in other dialog-based tasks~\cite{thomason2020vision,alamri2019audio,rashkin2018towards,li2017dailydialog} since they are based on relatively small-scaled datasets, and scaling up the size of the human-dialog datasets is laborious and expensive.

\noindent\textbf{Question type analysis.} We conduct a question-type analysis to identify what type of questions obtain benefits from GST. We divided the question type into six categories, \textit{Yes/No, Color, Objects, Counting, Time/Place,} and \textit{Others}. In Table~\hyperref[tab:qtype]{3}, the student model obtains more gains compared with the teacher model in less frequent question types (\textit{e.g.,} Counting and Time / Place). 

\subsection{Adversarial robustness results}
\label{sec:adv} We introduce a comprehensive evaluation setup for adversarial robustness in VisDial. Specifically, we propose three different adversarial attacks: (1) the FGSM attack, (2) a coreference attack, and (3) a random token attack. The FGSM attack perturbs input visual features, and the others attack the dialog history (\textit{i.e.,} textual inputs). \\

\noindent\textbf{Baselines.} We compare our student model against two ablative baselines: (1) the teacher model and (2) the student model utilizing the entire CC12M images without applying the in-domain image retrieval (\textit{i.e.,} student-iter1-full). We propose the student-iter1-full model to study the effect of the discarded images and the corresponding synthetic dialog data on adversarial robustness. \\
\begin{table}[ht!]
\centering
\vspace{-0.8cm}
\resizebox{0.48\textwidth}{!}{
\begin{tabular}{lccccccc}
\hline
\toprule
\multirow{3}{*}{Model} & 
\multicolumn{6}{c}{Question Type} \\
\cmidrule(lr){2-7} & Yes / No & Color & Objects & Counting & Time / Place  & Others \\
& (60.4\%) & (14.8\%) & (5.1\%) & (3.1\%) & (8.5\%) & (9.0\%) \\
\midrule
Teacher & 66.87 & 60.61 & 53.67 & 49.44 & 69.36 & 61.32 \\
\midrule
\textbf{Student} & \textbf{\makecell{67.41 \\ (\textcolor{blue}{+0.54})}} & \textbf{\makecell{61.85 \\ (\textcolor{blue}{+1.24})}} & \textbf{\makecell{55.25 \\ (\textcolor{blue}{+1.58})}} & \textbf{\makecell{51.76 \\ (\textcolor{blue}{+2.32})}} & \textbf{\makecell{71.38 \\ (\textcolor{blue}{+2.02})}} & \textbf{\makecell{63.02 \\ (\textcolor{blue}{+1.70})}} \\
\bottomrule
\hline
\end{tabular}
}
\caption{Question type analysis on the VisDial v1.0 validation split. The percentage denotes the data proportion of each category.}
\label{tab:qtype}
\end{table}

\begin{table*}[ht!]
\centering
\resizebox{0.8\textwidth}{!}{
\begin{tabular}{lcccccc}
\hline
\toprule
\multirow{2}{*}{Model} & 
\multirow{2}{*}{No Attack} &
\multirow{2}{*}{Coreference Attack} &
\multicolumn{4}{c}{Random Token Attack} \\
\cmidrule(lr){4-7} & & & 10\% & 20\% & 30\% & 40\% \\
\midrule
Teacher & 56.55 & 52.60 & 54.69\stdv{1.12} & 52.86\stdv{0.79} & 49.41\stdv{2.09} & 45.04\stdv{2.28} \\
Student (iter1, full) & 58.53 & 54.26 & 56.59\stdv{1.37} & 54.55\stdv{1.15} & 50.98\stdv{2.06} & 46.56\stdv{1.96} \\
\midrule
Student (iter1) & 58.63 & 54.34 & 55.59\stdv{0.88} & 54.26\stdv{1.54} & 51.04\stdv{2.39} & 47.04\stdv{2.03}  \\
Student (iter2) & 56.92 & 52.69 & 55.59\stdv{0.88} & 53.57\stdv{1.40} & 49.95\stdv{1.91} & 46.82\stdv{2.02} \\
\textbf{Student (iter3)} & \textbf{59.30} & \textbf{55.44} & \textbf{57.25}\stdv{0.91} & \textbf{55.10}\stdv{1.50} & \textbf{52.11}\stdv{2.75} & \textbf{48.00}\stdv{2.90} \\
\bottomrule
\hline
\end{tabular}
}
\caption{Adversarial robustness results against the attacks on the dialog history. We apply two different dialog history attacks: a coreference attack and a random token attack. The models are evaluated on the VisDialConv dataset~\cite{agarwal2020history} with the NDCG metric. The standard deviations are reported $\pm$ with five different random seeds.}
\label{tab:textattack}
\end{table*}

\noindent\textbf{Adversarial robustness against the FGSM attack.} The Fast Gradient Signed Method (FGSM) \cite{goodfellow2014explaining} is a white-box attack that perturbs the visual inputs based on the gradients of the loss with respect to the visual inputs. Formally, 
\begin{equation}
\begin{split}
    \mathrm{FGSM}(x) = x + \epsilon \cdot \mathrm{sign}(\nabla_{x}\mathcal{L}(x, y))
\end{split}    
\end{equation}
where $x$ and $y$ denote the visual inputs and the corresponding ground-truth labels, respectively. $\epsilon$ is a hyperparameter that adjusts the intensity of perturbations. However, different from the above setup, each question in VisDial can have one or more relevant answers in the list of answer candidates. We thus define the FGSM attack for VisDial as follows:
\begin{equation}
\label{eq:fgsm}
\begin{split}
    \mathrm{FGSM}(v) = v + \epsilon \cdot \mathrm{sign}(\sum_{c=1}^C r(a_{t,c}) \cdot  \nabla_{v}\mathcal{L}(c_t, a_{t,c}))
\end{split}
\end{equation}
where $C=100$ and $r(\cdot)$ denote the number of answer candidates and a function that returns the human-annotated relevance scores for each answer candidate, respectively. The relevance scores range from 0 to 1. $c_t$ and $a_{t,c}$ are the context triplet (\textit{i.e.,} $c_t\triangleq(v, d_{<t}, q_t)$) and the $c$-th answer candidate, respectively. The Equation~\hyperref[eq:fgsm]{4} indicates that the gradients of the loss for all relevant answers are considered for the FGSM attack.
\begin{figure}[t!]
\centering
\label{fig:imgattack}
\includegraphics[width=0.47\textwidth]{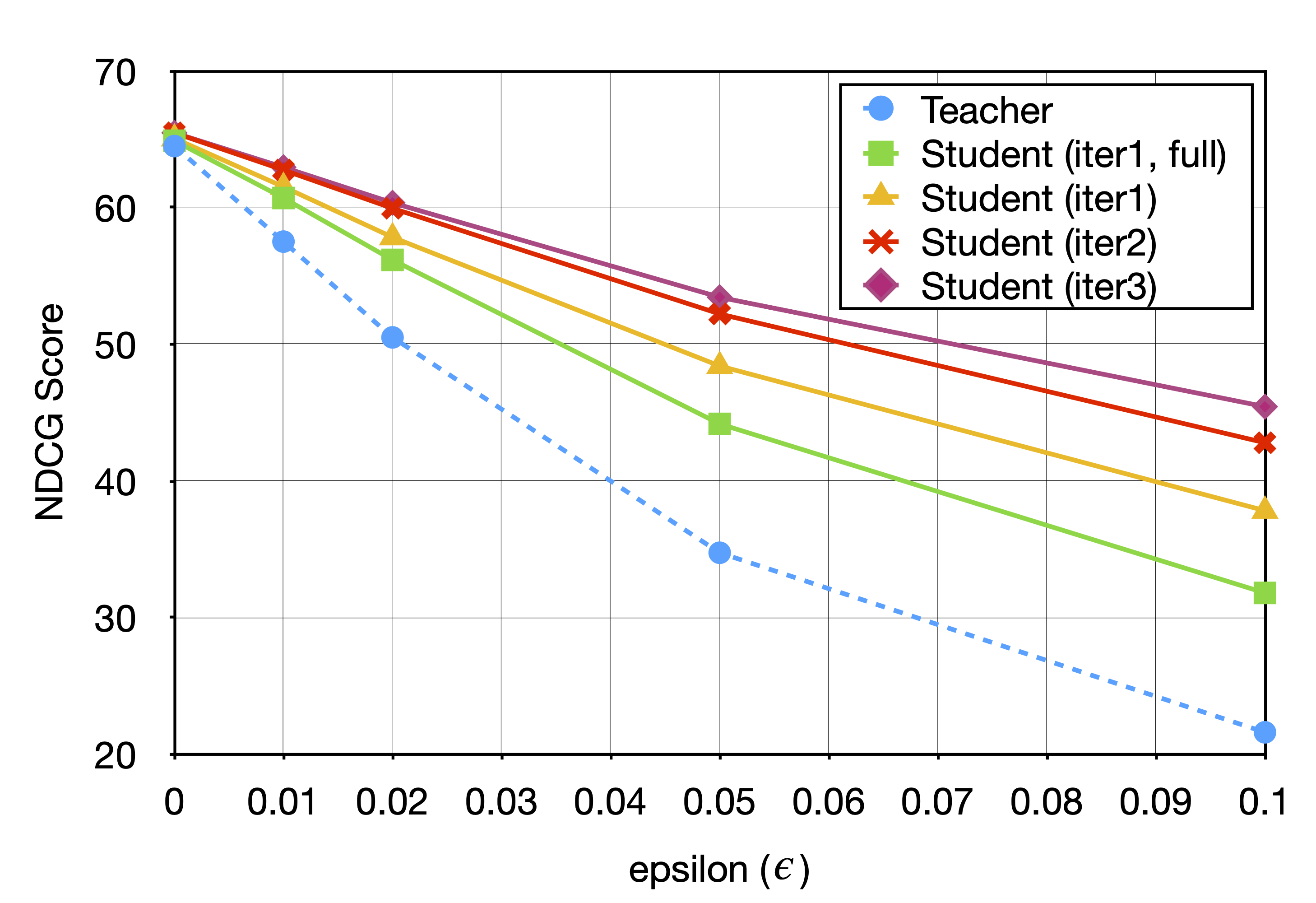}
\caption{Adversarial robustness against FGSM attack on VisDial v1.0 validation split. We report NDCG scores of each model.}
\end{figure}

As shown in Figure~\hyperref[fig:imgattack]{2}, we validate the models with four different epsilon values $\epsilon \in \left\{0.01, 0.02, 0.05, 0.1 \right\}$. The student model shows very significant improvements in NDCG compared with the teacher model. Specifically, the performance gap between the student model with three iterations (\textit{i.e.,} student-iter3) and the teacher model widens up to 23.83 absolute points (21.60 $\rightarrow$ 45.43) when $\epsilon$ is 0.1. It illustrates that GST makes the visual dialog model robust against the FGSM attack even though the student model is not optimized for adversarial robustness. Furthermore, we can clearly identify the efficacy of the iterative training as the intensity of the perturbations increases. The NDCG scores are boosted from 37.82\% (iter1) to 45.43\% (iter3) at $\epsilon=0.1$. Finally, the student-iter1 model shows better performance than the student-iter1-full model. It implies that the additional use of the discarded images along with the synthetic dialog does not bring any gains in the FGSM attack. \\

\noindent\textbf{Adversarial robustness against the textual attacks.} We also study the adversarial robustness against textual attacks to illustrate the effect of GST. We decide to perturb the dialog history because it contains useful information to answer the given question (\textit{e.g.,} cues for pronoun). However, according to recent studies~\cite{agarwal2020history,kang2020reasoning} in VisDial, not all questions require the dialog history to respond with the correct answers. So the work~\cite{agarwal2020history} has proposed a challenging subset of the VisDial validation dataset called VisDialConv. The VisDialConv dataset only contains questions that necessarily require the dialog history to answer (\textit{e.g.,} can you tell what it is for?). The crowd-workers conducted a manual inspection to select such \textit{context-dependent} questions.

Based on the VisDialConv dataset, we apply two different black-box attacks. First, we propose the coreference attack, which substitutes the noun phrases or pronouns in the dialog history with their synonyms to fool the VisDial models. Specifically, we leverage the off-the-shelf neural coreference resolution tool\footnote{\href{https://github.com/huggingface/neuralcoref}{https://github.com/huggingface/neuralcoref} based on the work \cite{clark2016deep}.} and find words in the dialog history that refer to objects such as those mentioned in a given question. We also borrow the counter-fitting word embeddings~\cite{mrkvsic2016counter} similar to textfooler~\cite{jin2020bert} to retrieve the synonyms. We greedily substitute the words with the synonyms with a minimum cosine distance in the embedding space since we observe that the other synonyms harm the original semantics of the dialog history. In Table~\hyperref[tab:textattack]{4}, the student-iter3 model outperforms the teacher model on NDCG by a large margin (2.84\%, 52.60 $\rightarrow$ 55.44) in the coreference attack. Furthermore, we do not see any merit in utilizing the entire CC12M~\cite{changpinyo2021conceptual} images and the corresponding synthetic dialog data, comparing the student-iter1-full with the student-iter1. 

The random token attack randomly replaces the word or sub-word tokens in the dialog history with a special \texttt{[MASK]} token. The pre-trained BERT$_{\texttt{BASE}}$ model~\cite{devlin2018bert} then recovers the masked tokens with masked language modeling (MLM) similar to BERT-ATTACK~\cite{li2020bert}. Finally, the perturbed dialog history is fed into the visual dialog models. We conduct this experiment by adjusting the probability of random masking up to 40\%. As shown in Table~\hyperref[tab:textattack]{4}, we evaluate each model with five random seeds and report the arithmetic mean and the standard deviations. The results demonstrate that GST is relatively robust against the random token attack compared with the baseline models.  

\begin{figure*}[btp]
\centering
\label{fig:vis}
\makebox[\textwidth]{\includegraphics[width=0.92\textwidth]{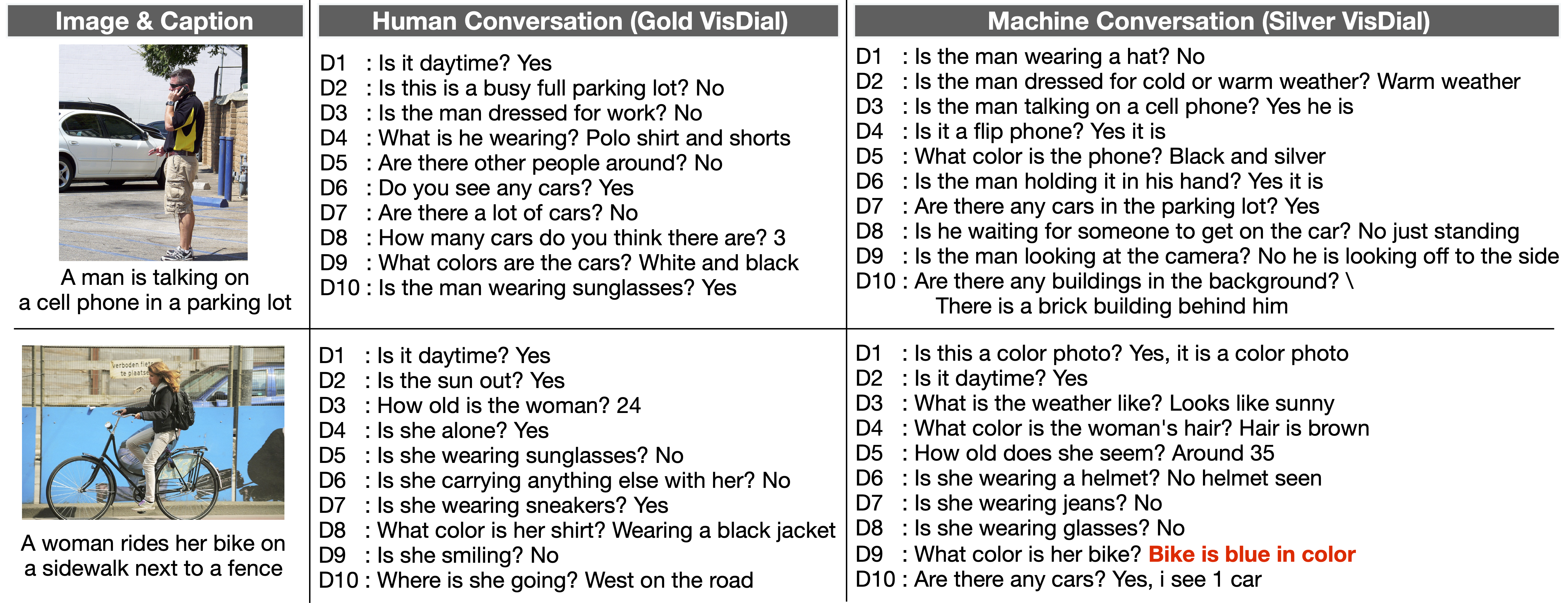}}
\caption{A visualization of the gold and the silver data on VisDial v1.0 validation split.}
\end{figure*}

\subsection{Analysis of the silver VisDial data}
\noindent\textbf{Comparison between silver and gold data.} For qualitative analysis of the silver data, we visualize the generated conversations from our proposed models and the ones from humans. We excerpt the human conversation from the VisDial v1.0 validation dataset, and the questioner and the student generate the machine conversation using the image and the caption in the validation data. As shown in Figure~\hyperref[fig:vis]{3}, diverse visual questions are generated in the silver VisDial data. For example, in D10 of the last example, the questioner asks about ``a car'' not mentioned by the human questioner and not even presented in the image caption. The student responds correctly to the question. Likewise, from D3 to D6 in the first example, the questioner deals with ``a cell phone,'' whereas the human questioner deals with different topics. However, we identify that the student sometimes fails to generate correct answers (\textit{i.e.,} the red-colored text), showing the importance of more precise visual grounding. \\ %Please see Appendix C for more examples.

\noindent\textbf{The diversity of silver questions.} We further quantify the generated question's diversity by comparing the gold questions with the silver ones for the same images in the VisDial v1.0 validation dataset. We extract N-grams for every ten questions (\textit{i.e.,} per image) in the gold and silver data and compare the N-grams between the two. We define the question diversity as the percentage of \textit{unique} silver N-grams not observed in the gold N-grams. We identify the question diversity by adjusting N from one to four. We generate three silver datasets and report the mean and standard deviations of the question diversity since the questioner performs stochastic decoding (see Appendix D). In Table~\hyperref[tab:qdiversity]{5}, the diversity significantly increases as N increases (92.80\% at N=4). It indicates that the questioner mainly generates different and distinctive 4-grams compared with the human questioner. \begin{table}[ht!]
\centering
\resizebox{0.45\textwidth}{!}{
\begin{tabular}{lccccc}
\hline
\toprule
\multirow{2}{*}{Model} & 
\multicolumn{4}{c}{N-gram Diversity} &  
\multirow{2}{*}{No Match} \\
\cmidrule(lr){2-5} & N=1 & N=2 & N=3 & N=4 & \\
\midrule
\textbf{Questioner} & \makecell{\textbf{28.06} \\ \stdv{0.14}} & \makecell{\textbf{56.46} \\ \stdv{0.09}} & \makecell{\textbf{76.98} \\ \stdv{0.08}} & \makecell{\textbf{92.80} \\ \stdv{0.08}} & \makecell{\textbf{95.38} \\ \stdv{0.15}} \\
\bottomrule
\hline
\end{tabular}
}
\caption{The N-gram diversity of the generated questions on the VisDial v1.0 validation images. The standard deviations are reported $\pm$ with three silver datasets. No match denotes the percentage of silver questions that do not precisely match the gold questions.}
\label{tab:qdiversity}
\end{table}
Furthermore, as shown in No Match at Table~\hyperref[tab:qdiversity]{5}, the questioner rarely generates the same questions that belong to gold questions. We analyze the answer diversity in Appendix C.

\subsection{Ablation study} The results of an ablation study are in Appendix B.2.

%\subsection{Qualitative analysis}
%We visualize the generated conversations from our models and the ones from humans. We excerpt the human conversation from the VisDial v1.0 validation set, and the questioner and the student generate the machine conversation using the image and the caption in the validation data. In Figure~\hyperref[fig:vis]{3}, diverse visual questions and correct answers are generated in the machine conversation. For instance, in D10, the questioner asks about ``a car,'' which the human questioner did not ever mention. The student responds correctly to the question. However, we also identify that the student sometimes fails to generate correct answers (\textit{i.e.,} the red-colored text), showing the importance of more precise visual grounding. We refer readers to Appendix C for further qualitative analysis.

\section{Conclusion}
We propose a semi-supervised learning approach for VisDial, called GST, that generates a synthetic visual dialog dataset for unlabeled Web images via multimodal conditional text generation. GST achieves the new state-of-the-art performance on the VisDial v1.0 and v0.9 datasets. Moreover, we demonstrate the efficacy of GST in low-data regime and adversarial robustness analysis. Finally, GST produces diverse dialogs compared with the human dialog. We believe the idea of GST is generally applicable to other multimodal generative domains and expect GST to open the door to leveraging unlabeled images in many visual QA tasks.

\noindent\textbf{Acknowledgements.} This work was supported by the SNU-NAVER Hyperscale AI Center and the Institute of Information \& Communications Technology Planning \& Evaluation (IITP) (2021-0-01343-GSAI/40\%, 2022-0-00953-PICA/30\%, 2022-0-00951-LBA/20\%, 2021-0-02068-AIHub/10\%) grant funded by the Korean government.

%our methods show strong performance gains when the human-labeled dialog data is scarce. Next, we propose a comprehensive evaluation setup for adversarial robustness in VisDial and validate GST is effective against visual and textual adversarial attacks. 
\clearpage

\clearpage

%%%%%%%%% REFERENCES
{\small
\bibliographystyle{ieee_fullname}
\bibliography{egbib}
}

\appendix
\clearpage
{\Large \noindent\textbf{Supplementary Materials}} \\

\noindent The supplementary materials are organized as: 
\begin{itemize}
  \item Section~\hyperref[appendixa]{A} shows a detailed model architecture. 
  \item Section~\hyperref[appendixb]{B} presents further quantitative analysis.
  \item Section~\hyperref[appendixc]{C} presents further qualitative analysis.
  \item Section~\hyperref[appendixd]{D} presents implementation details.
  \item Section~\hyperref[appendixe]{E} shows discussion, including limitations, future work, and ethical considerations.
\end{itemize} 
\vspace{0.05cm}

\section{Details of model architecture}
\label{appendixa}
\begin{figure*}[t!]
\centering
\label{fig:architecture}
\includegraphics[width=0.95\textwidth]{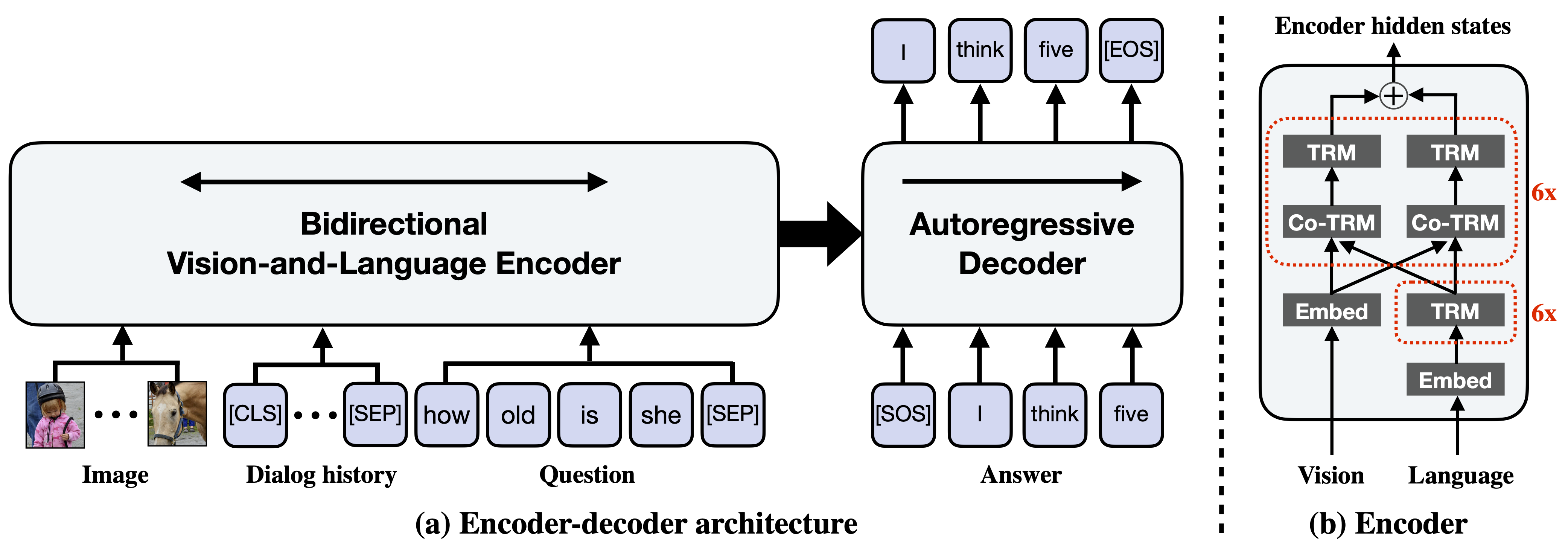}
\caption{A detailed architecture of our proposed model. We propose the encoder-decoder model where the encoder aggregates the given multimodal context, and the decoder generates the target sentence. (b): a more detailed view  of the encoder. TRM and Co-TRM denote the transformer module and the co-attentional transformer module, respectively. $\oplus$ denotes the concatenation operation.}
\end{figure*}

A detailed architecture of our proposed model is presented in Figure~\hyperref[fig:architecture]{4}. We use the encoder-decoder model, where the encoder aggregates the multimodal context, and the decoder generates the target sentence using the hidden states of the encoder. The answerer models (\textit{i.e.,} the student and the teacher) utilize the given image, the dialog history, and the question as the context. On the other hand, the questioner uses the given image and the dialog history as context to generate the question.

We employ the ViLBERT model~\cite{lu2019vilbert} as our encoder. We employ the BERT$_{\texttt{BASE}}$ model~\cite{devlin2018bert} for sequence generation~\cite{rothe2020leveraging} as our autoregressive decoder. The decoder has 12 layers of transformer blocks, with each block having 12 attention heads and a hidden size of 768. We present a detailed view of the encoder in (b) for Figure~\hyperref[fig:architecture]{4}. The encoder consists of the vision stream and the language stream. The language stream is the same model as the decoder (\textit{i.e.,} BERT$_{\texttt{BASE}}$), which has 12 layers of transformer blocks. The vision stream has 6 layers of transformer blocks, with each block having 8 attention heads with a hidden size of 1024. The co-attention layers connect the 6 transformer layers in the vision stream to the last 6 transformer layers in the language stream. The encoder concatenates the hidden states of each stream and passes them to the decoder. The decoder generates the target sentence by using them.

\section{Further quantitative analysis}
\subsection{Experiments on the discriminative models}
In this subsection, we discuss the details regarding GST for the discriminative visual dialog. We first describe how we can adapt GST to the discriminative models and then show the results on VisDial v1.0 test-standard split. \\ 

\label{appendixb}
\noindent\textbf{Model architecture.} Although our main focus is the generative model, we conduct additional experiments to identify the effect of GST in the discriminative VisDial model. Our proposed models (\textit{i.e.,} the student, the teacher, and the questioner) are based on encoder-decoder architecture where the encoder is based on the vision-and-language encoder model~\cite{lu2019vilbert}, and the decoder is the transformer decoder~\cite{rothe2020leveraging}. In this experiment, we remove the decoder model, so the student is based on the encoder-only architecture, the same model architecture as the ViLBERT model~\cite{lu2019vilbert}. We describe more details in the following subsection.\\

\begin{table*}
\centering
\resizebox{0.6\textwidth}{!}{
\begin{tabular}{lcccccc}
\hline
\toprule
\multirow{2}{*}{} & \multicolumn{6}{c}{VisDial v1.0 (test-std)}\\ 
\cmidrule(lr){2-7}
Model & NDCG$\uparrow$ & MRR$\uparrow$ & R@1$\uparrow$ & R@5$\uparrow$ & R@10$\uparrow$ & Mean$\downarrow$ \\
\midrule
CorefNMN~\cite{kottur2018visual} & 54.70 & 61.50 & 47.55 & 78.10 & 88.80 & 4.40 \\
RvA~\cite{niu2018recursive} & 55.59 & 63.03 & 49.03 & 80.40 & 89.83 & 4.18 \\ 
Synergistic~\cite{guo2019image} & 57.32 & 62.20 & 47.90 & 80.43 & 89.95 & 4.17 \\
ReDAN~\cite{gan2019multi} & 61.86 & 53.13 & 41.38 & 66.07 & 74.50 & 8.91 \\
DAN~\cite{kang2019dual} & 57.59 & 63.20 & 49.63 & 79.75 & 89.35 & 4.30 \\
FGA~\cite{schwartz2019factor} & 52.10 & 63.70 & 49.58 & 80.97 & 88.55 & 4.51 \\
VD-BERT~\cite{wang2020vd} & 59.96 & 65.44 & 51.63 & 82.23 & 90.68 & 3.90 \\
VisDial-BERT~\cite{murahari2019large} & \underline{63.87} & \underline{67.50} & \underline{53.85} & \underline{84.68} & \underline{93.25} & \underline{3.32} \\ 
\midrule
\textbf{Student (ours)} & \textbf{64.91} & \textbf{68.44} & \textbf{55.05} & \textbf{85.18} & \textbf{93.35} & \textbf{3.23} \\
\midrule\midrule
P1+P2$\dagger$~\cite{qi2020two} & 71.60 & 48.58 & 35.98 & 62.08 & 77.23 & 7.48 \\
MCA$\dagger$~\cite{agarwal2020history} & 72.47 & 37.68 & 20.67 & 56.67 & 72.12 & 8.89 \\
SGL+KT$\dagger$~\cite{kang2020reasoning} & 72.60 & \underline{58.01} & \underline{46.20} & \underline{71.01} & \underline{83.20} & \underline{5.85} \\
VD-BERT$\dagger$~\cite{wang2020vd} & \textbf{74.54} & 46.72 & 33.15 & 61.58 & 77.15 & 7.18 \\
UTC$\dagger$~\cite{chen2022utc} & 74.32 & 50.24 & 37.12 & 63.98 & 79.88 & 6.48 \\
VisDial-BERT$\dagger$~\cite{murahari2019large} & \underline{74.47} & 50.74 & 37.95 & 64.13 & 80.00 & 6.28 \\
\midrule
\textbf{Student$\dagger$ (ours)} & 71.76 & \textbf{68.09} & \textbf{55.18} & \textbf{83.68} & \textbf{91.93} & \textbf{3.57} \\
\bottomrule
\hline
\end{tabular}
}
\caption{Test-std performance of the discriminative model on the VisDial v1.0 dataset. $\uparrow$ indicates higher is better. $\downarrow$ indicates lower is better. $\dagger$ denotes the use of dense labels.}
\label{tab:t1}
\end{table*}

%\begin{figure*}[btp]
%\centering
%\label{fig:vis}
%\makebox[\textwidth]{\includegraphics[width=0.95\textwidth]{images/figure_visualization.png}}
%\caption{A visualization of the gold and the silver data on VisDial v1.0 validation split.}
%\end{figure*}

\noindent\textbf{Tricks for adapting to a discriminative task.} The goal of the discriminative task is to retrieve the ground-truth answer from a list of answer candidates. Therefore, it implies that the gold VisDial dataset~\cite{das2017visual} contains the pre-defined answer candidates for each question to train and evaluate the discriminative models. However, the silver VisDial dataset generated by our proposed models does not include the answer candidates since the dataset is generated to train the generative models that do not need the answer candidates. To circumvent this issue, GST first trains the student model for the generative task, \textit{i.e.,} the encoder-decoder model, on the silver VisDial data. Then, we extract the trained weights of the encoder in the student and initialize the encoder-only model with the weights. Finally, the encoder-only model is trained to retrieve the ground-truth answer from the list of answer candidates using the gold VisDial dataset. This trick circumvents the need for the answer candidates when training the silver VisDial data. \\

\noindent\textbf{Results on VisDial v1.0 test split.} We compare the student model with the state-of-the-art approaches in the discriminative task, consisting of VisDial-BERT~\cite{murahari2019large}, UTC~\cite{chen2022utc}, VD-BERT~\cite{wang2020vd}, SGL+KT~\cite{kang2020reasoning}, P1+P2~\cite{qi2020two}, MCA~\cite{agarwal2020history}, FGA~\cite{schwartz2019factor}, ReDAN~\cite{gan2019multi}, DAN~\cite{kang2019dual}, Synergistic~\cite{guo2019image}, RvA~\cite{niu2018recursive}, and CorefNMN~\cite{kottur2018visual}. As shown in the upper part of Table~\hyperref[tab:t1]{6}, GST outperforms the state-of-the-art approaches on all evaluation metrics in the VisDial v1.0 test-standard split. It is worth noticing that GST boosts NDCG 1.04\% (63.87 $\rightarrow$ 64.91) compared with the VisDial-BERT model, whose configuration is almost the same as the student except for the use of the silver VisDial data. Furthermore, recent studies finetune the discriminative VisDial models on the densely annotated labels\footnote{https://visualdialog.org/challenge/2019\#evaluation} in the validation dataset and evaluate the models on the test set to boost NDCG. The dense annotation finetuning yields considerable improvements on NDCG and counter-effect on other metrics (\textit{i.e.,} MRR, R@k, and Mean) due to the trade-off relationship~\cite{murahari2019large} between NDCG and the others. To mitigate such performance polarization, we follow the knowledge transfer technique in SGL+KT~\cite{kang2020reasoning} when using the dense labels. In the below part of Table~\hyperref[tab:t1]{6}, the student model still shows competitive performance on NDCG, maintaining powerful performance on other metrics. \\

\begin{table*}[ht!]
\centering
\resizebox{0.85\textwidth}{!}{
\begin{tabular}{lccccccccccc}
\hline
\toprule
\multirow{2}{*}{Model} & 
\multirow{2}{*}{PPL} &
\multirow{2}{*}{MCR} &
\multirow{2}{*}{IIR} &
\multirow{2}{*}{Iteration} &
\multicolumn{7}{c}{VisDial v1.0 (val)} \\
\cmidrule(lr){6-12} & & & & & & NDCG$\uparrow$ & MRR$\uparrow$ & R@1$\uparrow$ & R@5$\uparrow$ & R@10$\uparrow$ & Mean$\downarrow$ \\
\midrule
Teacher & & & & 0 & & 64.50 & 52.06 & 42.04 & 62.92 & 71.06 & 14.54 \\
Teacher (w/ CPT) & & & \checkmark & 0 & & 63.59 & 51.70 & 41.99 & 61.88 & 68.62 & 16.21 \\
\midrule
Student (iter1, w/o PPL) & & \checkmark & \checkmark & 1 & & 63.96 & 52.33 & 42.68 & 62.52 & 69.47 & 15.56  \\
Student (iter1, w/o MCR) & \checkmark  & & \checkmark & 1 & & 63.71 & 52.49 & 42.56 & 62.87 & 70.00 & 15.21  \\
Student (iter1, w/o IIR) & \checkmark & \checkmark & & 1 & & 64.57 & 52.33 & 42.10 & 63.46 & \textbf{71.54} & \textbf{14.31} \\
Student (iter1) & \checkmark & \checkmark & \checkmark & 1 & & 65.06 & 52.84 & 42.74 & \underline{63.66} & 71.30 & 14.60 \\
Student (iter2) & \checkmark & \checkmark & \checkmark & 2 & & \underline{65.46} & \underline{53.04} & \textbf{43.15} & 63.63 & 71.00 & 14.73 \\
Student (iter3) & \checkmark & \checkmark & \checkmark & 3 & & \textbf{65.47} & \textbf{53.19} & \underline{43.08} & \textbf{64.09} & \underline{71.51} & \underline{14.34} \\
\bottomrule
\hline
\end{tabular}
}
\caption{Ablation study on the VisDial v1.0 validation split. CPT denotes continued pre-training.}
\label{tab:abl}
\end{table*}

\noindent\textbf{Results on VisDial v1.0 validation split.} We also compare GST with the state-of-the-art vision-and-language pre-training model, BLIP~\cite{li2022blip}. The BLIP model is trained on the large-scale image-text datasets, such as Laion-400M~\cite{schuhmann2021laion}, CC12M~\cite{changpinyo2021conceptual}, CC3M~\cite{sharma2018conceptual}, COCO~\cite{lin2014microsoft}, Visual Genome~\cite{krishna2017visual}, and SBU captions~\cite{ordonez2011im2text}. Then, the model is finally fine-tuned on VisDial data. GST trains the student model on nearly 6.7M images, including 3.1M images (CC3M~\cite{sharma2018conceptual} and VQA~\cite{antol2015vqa}) to pretrain ViLBERT~\cite{lu2019vilbert} and 3.6M images filtered from CC12M~\cite{changpinyo2021conceptual} to generate and train synthetic dialog data. As shown in Table~\hyperref[tab:t2]{8}, GST shows competitive performance on the VisDial v1.0 validation split, outperforming BLIP on MRR. It is noticeable that the BLIP model utilizes nearly twenty times more images than GST. It indicates that GST is effective and sample-efficient.   
\begin{table}[]
    \centering
    \resizebox{0.42\textwidth}{!}{
    \begin{tabular}{lccc}
    \hline
    \toprule
    \multirow{2}{*}{Model} & 
    \multirow{2}{*}{Pre-train \# Images} &
    \multicolumn{2}{c}{VisDial v1.0 (val)} \\
    \cmidrule(lr){3-4} & & NDCG$\uparrow$ & MRR$\uparrow$ \\
    \midrule
    BLIP~\cite{li2022blip} & 129M & - & 69.41  \\
    \midrule
    \textbf{Student (ours)} & 6.7M & 65.92 & \textbf{69.51} \\
    \bottomrule
    \hline
    \end{tabular}
    }
    \caption{Comparison with BLIP~\cite{li2022blip} on the VisDial v1.0 validation split. The Pre-train \# Images denotes the number of utilized images before finetuning on the VisDial v1.0 data.}
    \label{tab:t2}
\end{table}
\subsection{Ablation study}
We perform an ablation study to illustrate the effect of each component in GST. We report the performance of four ablative models: student w/o PPL, student w/o MCR, student w/o IIR, and teacher w/ CPT. Student w/o PPL denotes the model that utilizes all generated QA pairs without applying the perplexity-based data selection. Student w/o MCR does not inject noises into the inputs of the student model. Student w/o IIR utilizes the entire CC12M~\cite{changpinyo2021conceptual} images to generate the silver VisDial data without applying in-domain image retrieval. It is the same model as the student-iter1-full in Section~\hyperref[sec:adv]{4.3}. Lastly, the teacher with continued pre-training (CPT) continues to perform pre-training with image-caption pairs in the silver VisDial data. CPT is proposed to identify the effect of utilizing additional vision-and-language data. Specifically, masked language modeling loss and masked image region loss are optimized by following ViLBERT~\cite{lu2019vilbert}. 

In Table~\hyperref[tab:abl]{7}, we observe all components (\textit{i.e.,} PPL, MCR, and IIR) play a significant role in boosting the performance. Notably, by comparing the student model with the student w/o IIR, we find that utilizing the entire Web images does not contribute to an accurate answer prediction. Moreover, we observe that CPT results in a considerable drop in performance. We conjecture that it is due to low-precision image captions in the CC12M dataset, as mentioned in the paper~\cite{changpinyo2021conceptual}. But the student still shows competitive performance even if it also utilizes the captions in the dialog history. Finally, the iterative training monotonically improves the performance, similar to the robustness results in Section~\hyperref[sec:adv]{4.3}.

\subsection{Do performance improvements come from a larger computational cost?}
It takes more computational costs to train the student model than to train the teacher model due to the silver VisDial data. Accordingly, we perform an analysis to prove that the performance improvements do not merely come from larger computational costs. The training time of the teacher model is about 1 day with one NVIDIA A100 GPU. It takes 5 days to train the student model with three iterations (\textit{i.e.,} iter3). Accordingly, we compare the ensemble of 5 teacher models with the student model with the iter3. We ensemble 5 teacher models with different weight initialization and average logits for 5 teacher models to predict the answer. The results are shown in Table~\hyperref[tab:cc]{9}. The student model outperforms the ensembles of 5 teacher models on both metrics. It indicates that the improvements from GST do not merely come from increased computational costs.
\begin{table}[]
    \centering
    \resizebox{0.35\textwidth}{!}{
    \begin{tabular}{lcc}
    \hline
    \toprule
    \multirow{2}{*}{} & \multicolumn{2}{c}{VisDial v1.0 (val)} \\ 
    \cmidrule(lr){2-3}
    Model & NDCG$\uparrow$ & MRR$\uparrow$ \\
    \midrule
    Teacher (single model) & 64.50 & 52.06 \\
    Teacher (5 ensembles) & 64.82 & 52.51 \\
    \midrule
    \textbf{Student (single model)} & \textbf{65.47} & \textbf{53.19} \\
    \bottomrule
    \hline
    \end{tabular}
    }
    \caption{Comparison between the student model with the ensemble of the five teacher models on balanced computational costs.}
    \label{tab:cc}
\end{table}

\begin{table}[]
    \centering
    \resizebox{0.28\textwidth}{!}{
    \begin{tabular}{lc}
    \hline
    \toprule
    Model & QA Utilization \\
    \midrule
    Student (iter1) & 32.52\% \\
    Student (iter2) & 39.06\% \\
    \textbf{Student (iter3)} & \textbf{46.40\%} \\
    \bottomrule
    \hline
    \end{tabular}
    }
    \caption{We define QA utilization as the proportion of utilized QA pairs in the silver VisDial data after applying perplexity-based data selection (\textit{i.e.,} PPL). The selection threshold $\tau$ is fixed at 50.}
    \label{tab:qau}
\end{table}

\subsection{The QA utilization across different iterations}
We identify how many QA pairs in the silver VisDial data are actually utilized after applying perplexity-based data selection (\textit{i.e.,} PPL). Accordingly, we define QA utilization as the proportion of utilized QA pairs in the silver VisDial data. The QA utilization across different iterations is shown in Table~\hyperref[tab:qau]{10}. We observe that the QA utilization increases as the iteration proceeds. It implies that the student model leverages more data as the iteration proceeds, and more importantly, the average perplexity of the generated answers gradually decreases. We argue that the drop of the answer perplexity is closely related to the student model being more confident and remaining low-entropy~\cite{grandvalet2004semi,sohn2020fixmatch}.      

\section{Further qualitative analysis}
\label{appendixc}
\subsection{More visualization of silver data}
\begin{figure*}[t!]
\centering
\label{fig:vis3}
\makebox[\textwidth]{\includegraphics[width=0.83\textwidth]{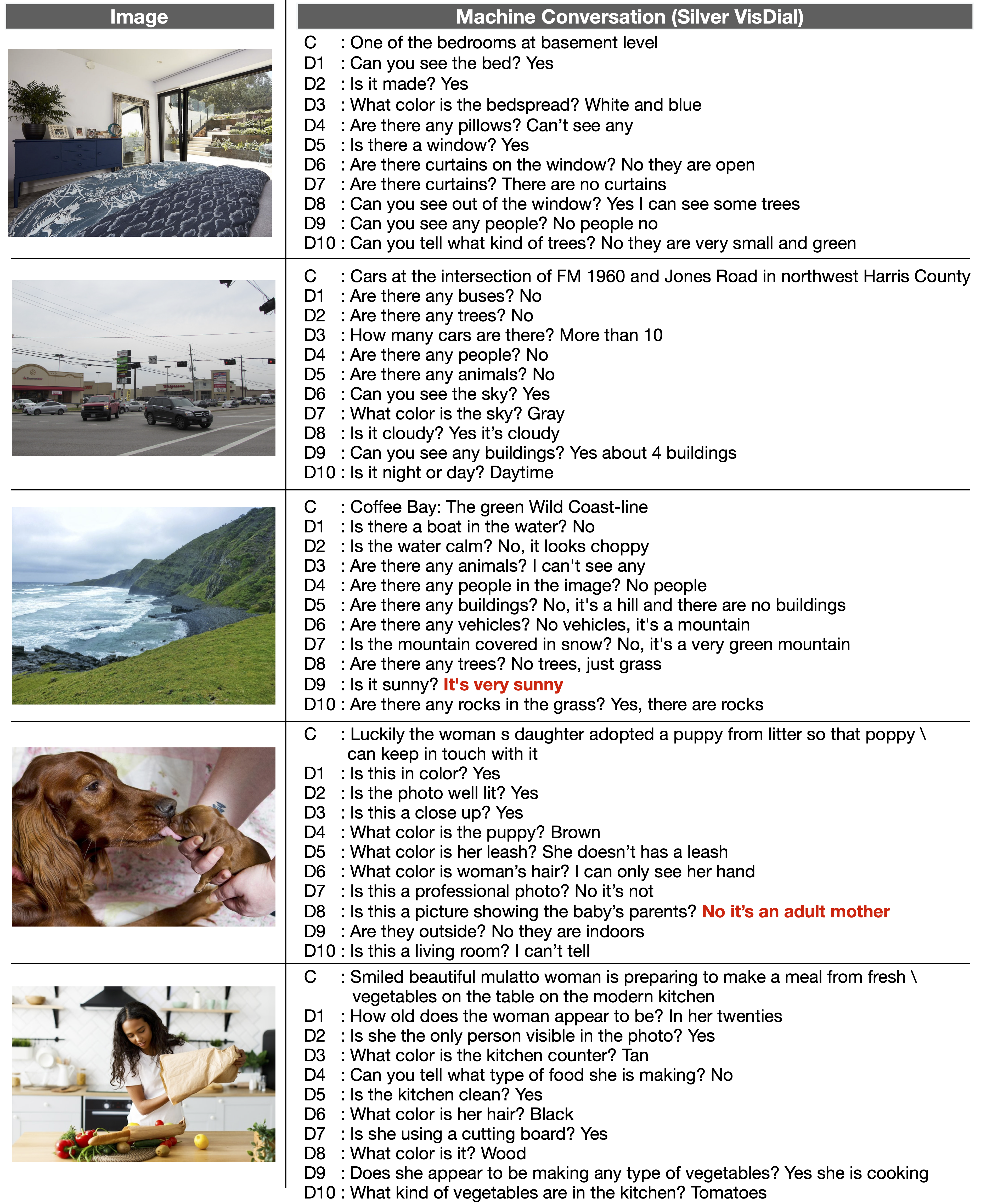}}
\caption{A visualization of the silver VisDial data based on the image-caption pairs in the Conceptual Captions 12M (CC12M)~\cite{changpinyo2021conceptual} dataset. The red-colored text denotes an incorrect answer.}
\end{figure*}

We visualize more silver data based on the image-caption pairs in the Conceptual Captions (CC12M)~\cite{changpinyo2021conceptual} dataset. As shown in Figure~\hyperref[fig:vis3]{6}, the questioner and the student models generate diverse and correct visual dialog data, although the image caption data is noisy. For instance, the image caption in the fourth example (\textit{i.e., Luckily the woman s daughter adopted a puppy from litter so that poppy can keep in touch with it}) is not well grounded with the given image. Still, our proposed models produce the \textit{visually-grounded} QA samples. Finally, the student sometimes fails to generate correct answers (the red-colored text), similar to Figure~\hyperref[fig:vis]{3}.

\subsection{Analysis of silver and gold answers.} 
\begin{figure*}[t!]
\centering
\label{fig:vis2}
\makebox[\textwidth]{\includegraphics[width=0.83\textwidth]{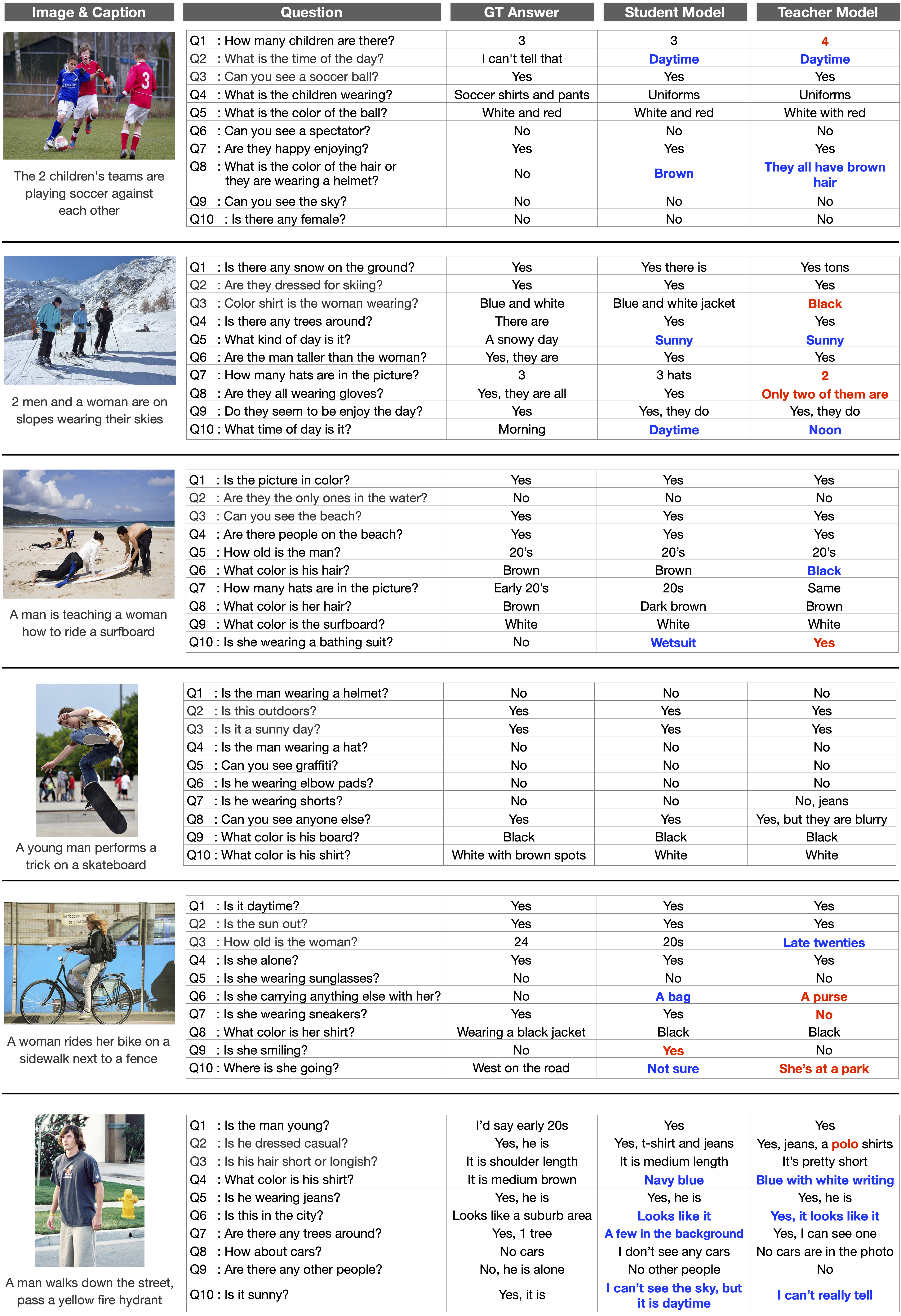}}
\caption{A visualization of answer predictions from the student and the teacher model. The red-colored text denotes an incorrect answer. The blue-colored text indicates the prediction different from the ground-truth answer, but it seems correct or plausible.}
\end{figure*}

We visualize the ground-truth answer (\textit{i.e.,} the gold answer) and the answer predictions from the student and the teacher models given the same context. As shown in Figure~\hyperref[fig:vis2]{5}, the student model indeed produces correct answers compared with the teacher model. Moreover, both models produce many correct or plausible answers, although the predicted answers differ from the gold answers (see the blue-colored text). For instance, for the last question in the third example (\textit{i.e., Is she wearing a bathing suit?}), the student answers ``wetsuit'' to the question, although the ground-truth answer is ``no''. We conjecture that the ability to generate such different yet correct answers is evaluated as a high NDCG performance; NDCG considers all relevant responses in the answer candidates. 

\section{Implementation details}
\label{appendixd}
We integrate the vision-and-language encoder~\cite{lu2019vilbert} with the transformer decoder for sequence generation (\textit{i.e.,} $\mathrm{BERT}_\text{BASE}$~\cite{rothe2020leveraging}) to train the teacher, the questioner, and the student. The decoder has 12 layers of transformer blocks, with each block having 12 attention heads and a hidden size of 768. The maximum sequence length of the encoder and the decoder is 256 and 25, respectively. We extract the feature vectors of the input images by using the Faster R-CNN~\cite{ren2015faster,Anderson2017up-down} pre-trained on Visual Genome~\cite{krishna2017visual}. The number of bounding boxes for each image is fixed to 36. We set the threshold for PPL $\tau$ to 50. We train on one A100 GPU with a batch size of 72 for 70 epochs. Training time takes about 3 days. We use the Adam optimizer~\cite{kingma2014adam} with an initial learning rate 1e-5. The learning rate is warmed up to 2e-5 until 10k iterations and linearly decays to 1e-5. In visually-grounded dialog generation, the questioner and the teacher decode the sequences using the top-$k$ sampling~\cite{fan2018hierarchical,holtzman2018learning,radford2019language} with $k=7$ and the temperature of 0.7. We use the top-$k$ sampling since its computation is cheap yielding accurate and diverse sequences. Furthermore, we apply the 4-gram penalty~\cite{paulus2017deep,klein2017opennmt} when generating visual questions to ensure that no 4-gram appears twice in the questions for each dialog.

\section{Discussion}
\label{appendixe}
\subsection{Relationship between self-supervised pre-training and generative self-training.} We develop the teacher, the questioner, and the student models on top of ViLBERT~\cite{lu2019vilbert} which leverages vision-and-language pre-training. Thus, the teacher can be understood as a typical model that follows the pretrain-then-transfer learning strategy mentioned in the introduction, whereas the student leverages both pre-training and generative self-training. By comparing the student with the teacher, we identify that self-supervised pre-training and GST are complementary modeling capabilities.

\subsection{Limitations and future work.} One of the major limitations of our approach is the learning efficiency of the student model. We demonstrate the effectiveness of our proposed method, but there can be more efficient ways to improve the visual dialog model. For example, our method generates the dialog data without considering the difficulty of the question. We believe that the competency-aware or curriculum-based visual dialog generation can make our proposed self-training algorithm more efficient and powerful. We will leave it as a future work.

\subsection{Ethical considerations.} Since GST generates the visually-grounded dialogs, our proposed models have the potential to produce biased and offensive language, although arguably to a lesser extent than the open-domain dialog~\cite{zhang2019dialogpt,shang2015neural,li2016deep,serban2017hierarchical,saleh2020hierarchical,li2017adversarial}. We attempt to mitigate ethical concerns such as biases against people of a certain gender, race, age, and ethnicity or the use of offensive content. Our proposed method utilizes the images and the captions in the Conceptual 12M dataset~\cite{changpinyo2021conceptual}, where several data cleansing processes (\textit{e.g.,} the offensive content filtering or replacing each person name with the special \texttt{<PERSON>} token) have been conducted. At least, we could not find any conversation violating the ethical considerations in a manual inspection by visualizing $\sim$100 synthetic dialogs. 
\vspace{1.5cm}

\end{document}